\documentclass{trbunofficial}
\makeatletter
\@ifundefined{Bbbk}{}{\let\Bbbk\@undefined}
\makeatother

\usepackage{xcolor}
\usepackage{multicol,multirow}
\usepackage[normalem]{ulem}
\usepackage{cleveref}
\usepackage{booktabs}
\usepackage{colortbl}
\usepackage{algpseudocode}
\newcounter{algorithm}
\usepackage{xcolor}
\usepackage{hyperref}

\newenvironment{algorithm}[1][!htbp]{%
    \par\smallskip\begingroup\refstepcounter{algorithm}%
    \noindent\rule{\linewidth}{0.6pt}\par\vspace{0.35em}%
    \noindent\textbf{Algorithm~\thealgorithm. }\ignorespaces
}{%
    \par\vspace{0.35em}\noindent\rule{\linewidth}{0.6pt}
    \par\smallskip\endgroup
}

\begin{document}


\title{From Vessel Trajectories to Safety-Critical Encounter Scenarios: A Generative AI Framework for Autonomous Ship Digital Testing}

\TRBauthor
  {Sijin Sun}
  {Senior Research Engineer, A*STAR IAIC}
  {sun\_sijin@a-star.edu.sg}
  [Singapore]

\TRBauthor*
  {Liangbin Zhao}
  {Senior Scientist, A*STAR IAIC}
  {zhao\_liangbin@a-star.edu.sg}
  [Singapore]

\TRBauthor
  {Xiuju Fu}
  {Senior Principal Scientist, A*STAR IAIC}
  {fuxj@a-star.edu.sg}
  [Singapore]

\AuthorHeaders{Sun, Zhao, Deng, and Fu}

\maketitle
\pagestyle{plain}

\section{Abstract}
\noindent\textbf{Objectives:} Safety validation of Maritime Autonomous Surface Ships (MASS) in constrained waterways requires diverse safety-critical encounters, yet these events are rare in historical Automatic Identification System (AIS) data, limiting comprehensive and systematic simulation testing. This study develops a framework that learns traffic-flow distributions and converts single-vessel trajectories into configurable crossing, head-on, and overtaking encounters with controllable timing and risk.

\hfill\break%
\noindent\textbf{Methods:} One year of AIS trajectories from the Singapore Strait was analyzed in two stages. First, a GeoAIS variational autoencoder (GAVAE) learned route-conditioned motion distributions using spatiotemporal features, vessel-motion embeddings, batch-level traffic context, dual-path decoding, and Route Calibration to generate geographically consistent, diverse, and kinematically plausible trajectories. Second, same- or different-route trajectories were paired under configurable temporal offsets and screened by overlap regions, minimum observed separation, dynamic DCPA/TCPA criteria, and encounter-type consistency. Retained encounters were standardized into pre-encounter, encounter, and post-encounter segments with vessel states and risk indicators for replay and evaluation.

\hfill\break%
\noindent\textbf{Findings:} Comparative evaluation showed GAVAE performed best overall among evaluated models in preserving route geometry, spatial variability, and traffic distributions. Generated pools produced diverse crossing, head-on, and overtaking encounters with relative-course-angle, DCPA, and TCPA distributions matching intended geometries and smooth motion across three stages. Sensitivity analysis showed temporal offsets and CPA thresholds controlled timing and risk, validating realistic, diverse, and simulation-ready scenarios.

\hfill\break%
\noindent\textbf{Novelty:} This AIS-driven maritime safety framework links route-conditioned trajectory generation with configurable pairing, temporal alignment, encounter type, and CPA-based risk controls. To our knowledge, it is the first to represent crossing, head-on, and overtaking encounters as parameterized, standardized three-stage scenarios for simulation-based MASS testing.

\hfill\break%
\noindent\textbf{Practical Applications:} The framework enables MASS developers, operators, and regulators to build scalable scenario libraries for pre-deployment evaluation of collision avoidance, recovery, and decision-support systems. Implementation requires representative high-quality local AIS data, route calibration, and temporal-offset and CPA thresholds so scenarios reflect local traffic context and regulatory risk criteria.

\newpage

\section{Introduction}
 Digital testing is a core component of Maritime Autonomous Surface Ship (MASS) development and evaluation, and the realism and effectiveness of test scenarios are critical to the credibility of the testing process. However, constructing scenarios that faithfully reflect real-world navigational risks remains a significant challenge. Scenarios manually configured based on expert knowledge may not fully capture both the complexity and difficulty of actual vessel encounters. Consequently, the use of historically observed encounters as the basis for scenario construction has received increasing attention.

Nevertheless, realistic safety-critical vessel encounter scenarios are particularly scarce in real-world maritime traffic. Although large-scale Automatic Identification System (AIS) datasets contain extensive vessel movement records, the vast majority represent routine navigation, while near-miss events and high-risk interactions occur only rarely. Therefore, direct extraction from historical AIS data often provides insufficient diversity and coverage for comprehensive MASS development and evaluation. An effective scenario set should preserve realistic vessel motion while covering diverse safety-critical conditions and plausible distributions of key encounter parameters, including the distance at the closest point of approach (DCPA), time to the closest point of approach (TCPA), and relative course angle.\cite{khastgir2021systems,neurohr2020fundamental}.

Existing maritime scenario-generation studies address different aspects of scenario realism, controllability, and coverage. Knowledge-driven methods construct COLREGs-based or template-based encounters, providing explicit labels, repeatable geometries, and controllable vessel motion patterns \cite{porres2020scenariobased,sawada2024framework}. Formal scenario specifications further improve traceability by representing encounters through predefined abstractions and constraints \cite{austel2024formal}. AIS-based extraction and random-sampling approaches improve realism by deriving encounter configurations or parameter distributions from observed vessel traffic \cite{zhu2022randomly,wang2024ship}, while navigation-oriented sampling organizes historical trajectories into representative encounter contexts \cite{hwang2021navigation}. Parameter-space exploration and simulation-driven approaches expand coverage by systematically varying encounter parameters \cite{bolbot2022automatic,zhu2025highrisk}. Structured data-driven pipelines further enhance controllability by describing encounters through abstract variables and predefined construction procedures \cite{rong2026datadriven}.

Despite these advances, existing methods face a fundamental trade-off between realism and scenario diversity. Knowledge-driven and parameterized approaches provide strong controllability, but their vessel motions and parameter combinations may oversimplify real navigation behavior or fail to follow the joint distributions observed in actual traffic. In contrast, AIS extraction, random sampling, and trajectory recombination preserve empirical realism, but the resulting scenarios remain bounded by the trajectories and encounters already present in the original dataset. Rare safety-critical conditions therefore remain underrepresented, even when large volumes of historical AIS data are available.

A gap remains between generating realistic individual trajectories and building valid safety-critical multi-vessel encounters. Route-specific trajectory pools support pairing and temporal alignment, but trajectory recombination alone cannot distinguish incidental spatial proximity from dynamically converging encounters. A construction method is therefore needed to combine generated trajectories within a shared encounter region, evaluate both observed separation and relative-motion CPA conditions, and export standardized temporal segments for simulator replay.

To address these gaps, this study proposes a two-stage framework \footnote{Source code and demo videos of the generated scenarios are available at:
\href{https://github.com/StanleySun233/gavae-scenario-construction}
{\fcolorbox{green!60!black}{white}{\textcolor{green!50!black}{GAVAE}}}.} for generating realistic safety-critical encounters from AIS data. In the first stage, GAVAE learns route-conditioned vessel motion distributions and generates geographically consistent trajectory variations by combining spatiotemporal AIS representations, maritime motion features, latent traffic patterns, and route-aware calibration. In the second stage, generated trajectories are paired across traffic flows for crossing and head-on encounters, while overtaking encounters use two different trajectories from the same flow. Temporal offsets align the candidate pairs, and region, proximity, relative-motion, encounter-type consistency, and complete-window conditions identify safety-critical scenarios centered on the closest interaction.

The main contributions of this work are summarized as follows:

\begin{itemize}

    \item Proposes a two-stage, data-driven framework that links route-conditioned trajectory generation with configurable encounter construction, transforming historical AIS data into standardized and simulation-ready safety-critical scenarios for MASS testing.

    \item Develops GAVAE, a maritime-aware variational autoencoder that integrates spatiotemporal AIS features, vessel-motion embeddings, batch-level traffic context, dual-path trajectory decoding, and route calibration to generate geographically consistent, diverse, and kinematically plausible single-vessel trajectories.

    \item Establishes a structured encounter-construction process that combines inter- and intra-route trajectory pairing, temporal alignment, data-derived overlap-region constraints, observed-distance and dynamic CPA screening, and encounter-type verification. Each retained case is represented through standardized pre-encounter, encounter, and post-encounter segments for simulator replay and evaluation.

\end{itemize}

\section{Literature Review}
\subsection{Safety-Critical Vessel Encounter Scenario Generation}

The construction of safety-critical encounter scenarios has attracted increasing attention in recent years, particularly for simulation-based testing of autonomous navigation and decision-support systems. Scenario-based testing studies in automated transport define test value through scenario relevance and coverage \cite{khastgir2021systems,neurohr2020fundamental}. General scenario-definition work also frames scenarios through relevant system states \cite{ulbrich2015defining}. In the maritime domain, existing approaches to scenario generation can be broadly categorized into knowledge-driven, data-statistical, and generative methods, each addressing different aspects of this coverage problem.

Knowledge-driven approaches construct encounter scenarios from navigation rules, expert knowledge, or predefined geometric configurations. COLREGs-based templates make head-on, crossing, and overtaking cases reproducible and interpretable \cite{porres2020scenariobased,sawada2024framework}. Formal specifications organize scenario sets around regulatory coverage \cite{austel2024formal}. Their spatial variability follows route geometry and vessel kinematics specified by the scenario designer.

Data-statistical approaches reconstruct or sample encounters from historical AIS records and navigation parameters \cite{zhu2022randomly,wang2024ship}. AIS mining also organizes navigation contexts and traffic conflicts for testing \cite{hwang2021navigation,lei2020mining}. These methods preserve observed traffic patterns but inherit the sparsity of historical safety-critical events.

Algorithmic and generative approaches expand the scenario space beyond historical replay. Sampling in the encounter parameter space searches for hazardous traffic situations \cite{bolbot2022automatic}, while data-driven pipelines vary traffic configurations under explicit construction logic \cite{rong2026datadriven}. Learning-based search further targets high-risk interactions through system feedback \cite{zhu2025highrisk}. Motion realism and encounter test logic are commonly coupled inside the generator or represented through abstract encounter parameters.

Overall, existing approaches address complementary aspects of scenario generation, including interpretability, realism, and controllability. However, a systematic mechanism that can jointly preserve real traffic characteristics while enabling scalable generation of diverse safety-critical encounter scenarios remains limited. Motivated by these limitations, the proposed framework aims to preserve the realism of historical traffic data while expanding the availability of safety-critical scenarios through generative trajectory modeling.

\subsection{Vessel Trajectory Generation}
A scenario can be interpreted as a temporal sequence of system states, which may be represented using different modalities depending on the level of abstraction required for testing and analysis \cite{ulbrich2015defining}. For many traffic applications, trajectories provide a compact and effective representation of object motion, making them widely used in scenario modeling and simulation studies.

AIS trajectories support traffic analysis, route extraction, behavior modeling, and collision risk assessment \cite{lee2022maritime,lei2020mining}. Trajectory prediction estimates future vessel positions from historical motion patterns \cite{park2021ship}. These methods support navigation through future-position estimation, while systematic testing requires diverse motion generation.

Autoencoder and latent-variable models learn compressed motion distributions from historical trajectories \cite{murray2020dual}, while deep generative trajectory models learn synthetic vessel motion with interaction structure \cite{zhu2024deep}. Generative adversarial networks and variational autoencoders provide distribution-learning mechanisms for sequence synthesis \cite{goodfellow2014generative,kingma2013autoencoding}. These models support learned trajectory variation.

Motivated by these developments, the present work adopts a variational autoencoder to generate route-conditioned vessel trajectory variations for the subsequent construction of safety-critical encounter scenarios.

\section{Methodology}
\subsection{Problem Definition and Pipeline}

\begin{figure*}[!htbp]
\centering
\includegraphics[width=1.0\linewidth]{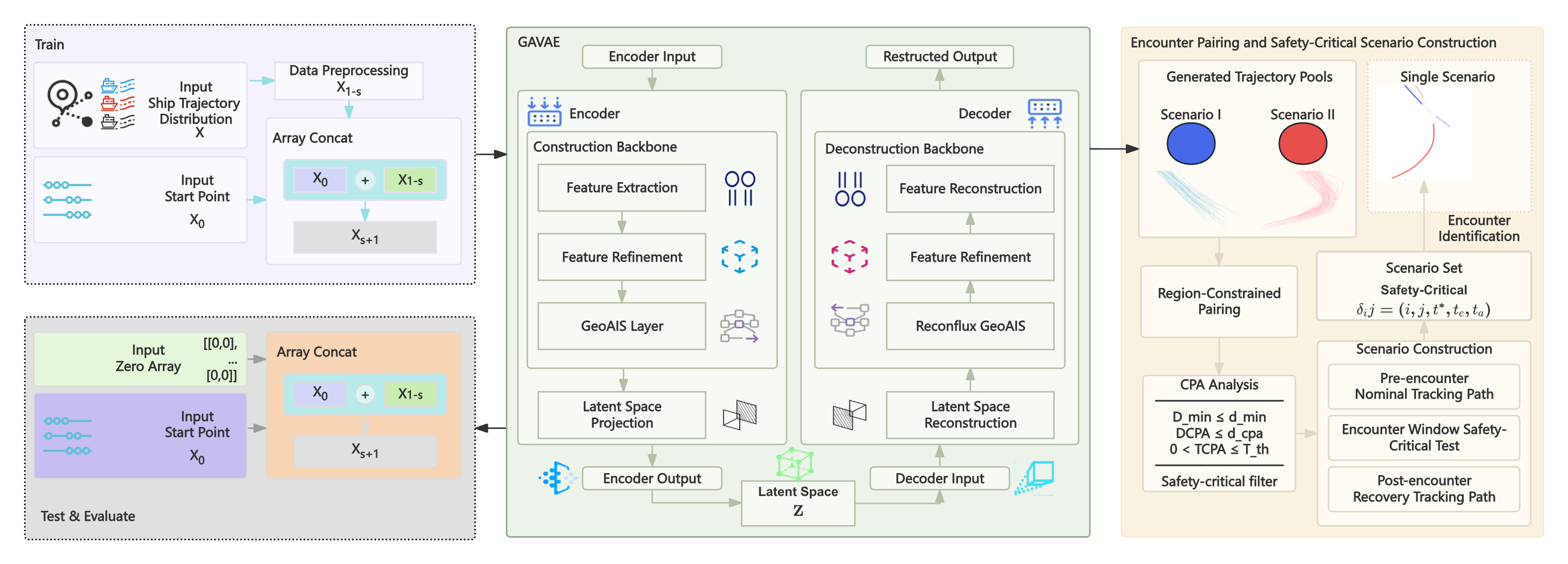}
\caption{Overview of the trajectory generation and scenario construction pipeline.}
\label{fig:model_structure}
\end{figure*}

The objective is to transform AIS trajectories from designated traffic flows into a set of structured encounter cases for simulator replay, where each scenario is treated as a temporal sequence of vessel states \cite{ulbrich2015defining}. For route $r$, the processed input is a set of trajectories $\mathcal{X}^{r}=\{x_i^r\}_{i=1}^{N_r}$, where $x_i^r\in\mathbb{R}^{T_r\times2}$ stores longitude and latitude after route filtering, uniform resampling, interpolation, and sequence-length standardization. The trajectories are partitioned by MMSI into mutually exclusive training, validation, and test sets, $\mathcal{X}^{r}=\mathcal{X}_{\mathrm{tr}}^{r}\mathbin{\dot{\cup}}\mathcal{X}_{\mathrm{val}}^{r}\mathbin{\dot{\cup}}\mathcal{X}_{\mathrm{te}}^{r}$. The route generator and scenario constructor form a compositional mapping,
\begin{equation}
\hat{\mathcal{T}}^r_\theta
=\mathcal{C}_\rho\!\left(\left\{\operatorname{InverseScale}_{r}\!\left(D_\theta(z_k^r)\right)\right\}_{k=1}^{M_r},\mathcal{X}_{\mathrm{tr}}^{r}\right),
\quad
\mathcal{S}_\eta
=\Phi_\eta\!\left(\hat{\mathcal{T}}^{(1)}_\theta,\hat{\mathcal{T}}^{(2)}_\theta\right),
\end{equation}
where $D_\theta$ is the learned AIS Spatiotemporal Decoder, $\operatorname{InverseScale}_{r}$ reverses normalization fitted on $\mathcal{X}_{\mathrm{tr}}^{r}$, $M_r$ is the number of generated trajectories for route $r$, and $\mathcal{C}_\rho(\cdot)$ is Route Calibration with blend factor $\rho$. The encounter constructor $\Phi_\eta(\cdot)$ is parameterized by temporal offsets, an overlap region, observed-distance and dynamic CPA thresholds, and two temporal margins. Each retained scenario is represented as
\begin{equation}
\sigma_{ij}=\left(i,j,t^\star,t_{\mathrm{early}},t_{\mathrm{after}}\right),
\end{equation}
where $i$ and $j$ identify the paired trajectories, $t^\star$ is the closest-interaction instant, and $t_{\mathrm{early}}$ and $t_{\mathrm{after}}$ define the encounter interval around $t^\star$. The scenario also stores the pre-encounter, encounter, and post-encounter trajectory clips for both vessels.

The route generator and encounter constructor form separate stages. The generator learns an individual vessel route distribution $p_\theta(x^r)$ from historical AIS data. The constructor combines trajectories, applies temporal offsets, and evaluates observed separation and dynamic CPA indicators after temporal alignment.

\Cref{fig:model_structure} summarizes the pipeline. The AIS Spatiotemporal Encoder and Decoder learn vessel motion within each traffic flow, and Route Calibration forms the Generated Trajectory Pools. The scenario-construction stage applies Region-Constrained Pairing, CPA Analysis, and a Safety-Critical Filter before dividing retained trajectories into pre-encounter, encounter, and post-encounter segments. The retained cases form the Safety-Critical Scenario Set.

\begin{algorithm}[!htbp]
GAVAE-based MASS encounter scenario construction
\label{alg:gavae_scenario_pipeline}
\par\smallskip
\small
\begin{algorithmic}[1]
\Require $\{(\mathcal{X}_{\mathrm{tr}}^{r},\mathcal{X}_{\mathrm{val}}^{r})\}_{r=1}^{R}$, temporal offsets $\Delta$, thresholds $d_{\min},d_{\mathrm{th}},d_{\mathrm{cpa}},T_{\mathrm{th}}$, margins $t_{\mathrm{early}},t_{\mathrm{after}}$, overtaking-route index $r_O\in\{1,\ldots,R\}$.
\Ensure $\mathcal{S}_\eta$
\State $\mathcal{S}_\eta\gets\varnothing$
\For{$r=1,\ldots,R$}
    \State $(X_{\mathrm{tr}}^r,X_{\mathrm{val}}^r)\gets\operatorname{Normalize}_{\mathcal{X}_{\mathrm{tr}}^r}(\mathcal{X}_{\mathrm{tr}}^r,\mathcal{X}_{\mathrm{val}}^r)$
    \State $\Phi_{\mathrm{tr}}^r\gets\operatorname{VesselMotionEmbedding}(X_{\mathrm{tr}}^r)$
    \State $(\theta_r,\phi_r)\gets\operatorname{TrainGAVAE}(X_{\mathrm{tr}}^r,\Phi_{\mathrm{tr}}^r;X_{\mathrm{val}}^r,\Cref{eq:gavae_loss})$
    \State $\{(\mu_i^r,\ell_i^r)\}_{i=1}^{|\mathcal{X}_{\mathrm{tr}}^r|}\gets E_{\phi_r}(X_{\mathrm{tr}}^r)$
    \State $i_k\sim\operatorname{Uniform}\{1,\ldots,|\mathcal{X}_{\mathrm{tr}}^r|\}$,\quad $\epsilon_k\sim\mathcal{N}(0,I)$
    \State $Z^r\gets\{\mu_{i_k}^r+\exp(\ell_{i_k}^r/2)\odot\epsilon_k\}_{k=1}^{M_r}$
    \State $\hat{\mathcal{T}}_\theta^r\gets\mathcal{C}_\rho(\operatorname{InverseScale}_{r}(D_{\theta_r}(Z^r)),\mathcal{X}_{\mathrm{tr}}^{r})$
\EndFor
\State $\mathcal{P}_{C}\gets\hat{\mathcal{T}}_\theta^{(1)}\times\hat{\mathcal{T}}_\theta^{(2)}$
\State $\mathcal{P}_{H}\gets\hat{\mathcal{T}}_\theta^{(1)}\times\hat{\mathcal{T}}_\theta^{(2)}$
\State $\mathcal{P}_{O}\gets\{(\tau_i^{(r_O)},\tau_j^{(r_O)}):\tau_i^{(r_O)},\tau_j^{(r_O)}\in\hat{\mathcal{T}}_\theta^{(r_O)},i\neq j\}$
\For{$c\in\{C,H,O\}$}
    \State $\mathcal{R}_c\gets\operatorname{OverlapRegion}(\mathcal{P}_c)$
    \For{$(\tau_i,\tau_j)\in\mathcal{P}_c,\ \delta\in\Delta$}
        \State $(A_\delta,B_\delta)\gets\operatorname{Align}_{\delta}(\operatorname{ProjectNM}(\tau_i,\tau_j))$
        \State $D_r(k)\gets\|A_\delta(k)-B_\delta(k)\|_2$
        \State $k^\star\gets\arg\min_k D_r(k)$,\quad $D_{\min}\gets D_r(k^\star)$
        \State $(\mathrm{TCPA}(k),\mathrm{DCPA}(k))\gets\operatorname{DynamicCPA}(A_\delta,B_\delta)$
        \State $q\gets\exists k:\ A_\delta(k),B_\delta(k)\in\mathcal{R}_c,\ D_r(k)\leq d_{\mathrm{th}},\ 0<\mathrm{TCPA}(k)\leq T_{\mathrm{th}},\ \mathrm{DCPA}(k)\leq d_{\mathrm{cpa}}$
        \If{$D_{\min}\leq d_{\min}$ and $q$ and $\operatorname{TypeConsistent}(c,A_\delta,B_\delta)$ and $\operatorname{CompleteWindow}(k^\star,t_{\mathrm{early}},t_{\mathrm{after}})$}
            \State $t^\star\gets\operatorname{SourceTime}(\tau_i,\delta,k^\star)$
            \State $\sigma_{ij}\gets(i,j,t^\star,t_{\mathrm{early}},t_{\mathrm{after}})$
            \State $\mathcal{S}_\eta\gets\mathcal{S}_\eta\cup\{(\sigma_{ij},\operatorname{Segment}(\tau_i,\tau_j,\delta,k^\star,t_{\mathrm{early}},t_{\mathrm{after}}))\}$
        \EndIf
    \EndFor
\EndFor
\end{algorithmic}
\end{algorithm}

\subsection{GeoAIS VAE for Maritime Trajectory Generation}

GAVAE contains an AIS Spatiotemporal Encoder, a Maritime-Aware Latent Space, and an AIS Spatiotemporal Decoder. Its Maritime Domain Feature Layer comprises Vessel Motion Embedding, Hidden Traffic Knowledge Extraction, and Route Calibration.

The Vessel Motion Embedding exposes normalized route time and local coordinate variation before spatiotemporal encoding. Longitude and latitude are normalized separately before embedding calculation. For a normalized trajectory $x=\{\tilde{p}_t\}_{t=1}^{T}$ with $\tilde{p}_t=(\tilde{\lambda}_t,\tilde{\phi}_t)$, the 15-dimensional embedding is
\begin{equation}
\label{eq:vessel_motion_embedding}
\begin{aligned}
\varphi_t
&=\left[\tilde{p}_t,\;\tilde{p}_t-\tilde{p}_1,\;\Delta\tilde{p}_t,\;m_t,\;u_t,\;\Delta^2\tilde{p}_t,\;r_t,\;\sin\omega_t,\;\cos\omega_t,\;q_t\right],\\
\Delta\tilde{p}_t&=\tilde{p}_t-\tilde{p}_{t-1},\quad
m_t=\|\Delta\tilde{p}_t\|_2,\quad
u_t=\frac{\Delta\tilde{p}_t}{\max(m_t,\varepsilon)},\\
\Delta^2\tilde{p}_t&=\Delta\tilde{p}_t-\Delta\tilde{p}_{t-1},\quad
r_t=\|\Delta^2\tilde{p}_t\|_2,\quad
q_t=\frac{t-1}{T-1},\\
\sin\omega_t&=u_{t,x}u_{t-1,y}-u_{t,y}u_{t-1,x},\quad
\cos\omega_t=u_t^\top u_{t-1}.
\end{aligned}
\end{equation}
The Vessel Motion Embedding contains normalized position, offset from the first normalized point, first and second coordinate differences and their magnitudes, direction and angular change in scaled coordinate space, and normalized route time. The variables $m_t$, $r_t$, and $q_t$ represent first-difference magnitude, second-difference magnitude, and normalized route time.

\begin{figure*}[!htbp]
\centering
\includegraphics[width=.85\textwidth]{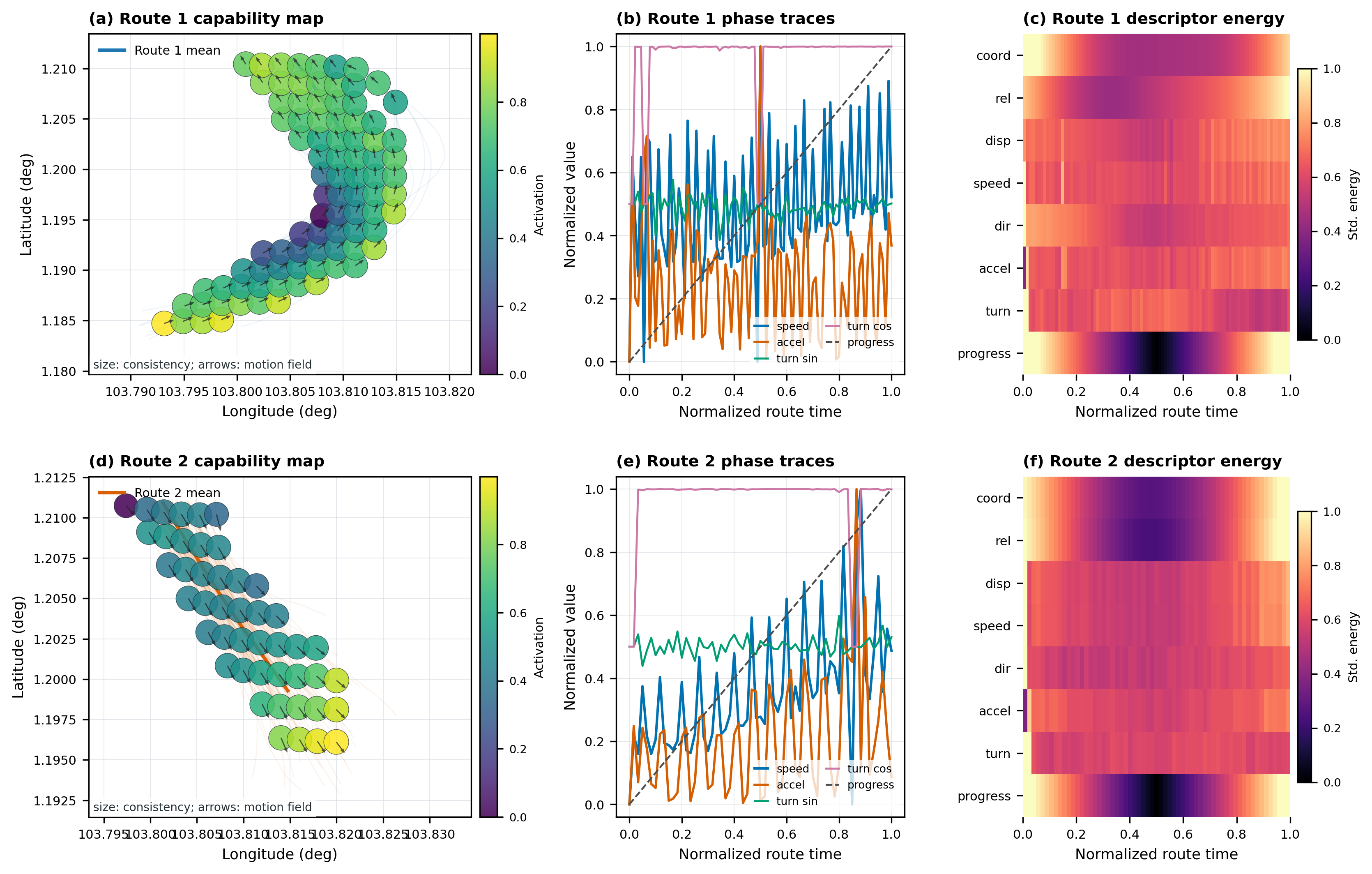}
\caption{Route-wise GeoAIS capability maps, phase traces, and descriptor energy.}
\label{fig:geoais_diagnostics}
\end{figure*}

The same embedding supports route-wise GeoAIS diagnostic visualization. Each route corridor is discretized separately into geographic cells, preserving opposing traffic directions in distinct coordinate fields. For cell $g$ with observations $\mathcal{I}_g=\{(i,t):p_{it}\in g\}$, motion-field, activation, and directional-consistency quantities are
\begin{equation}
\label{eq:geoais_diagnostics}
\begin{aligned}
\mathbf{f}_g&=\frac{1}{|\mathcal{I}_g|}\sum_{(i,t)\in\mathcal{I}_g}u_{it},\\
A_g&=\operatorname{scale}_{[0,1]}\!\left(
\frac{1}{|\mathcal{I}_g|}\sum_{(i,t)\in\mathcal{I}_g}(m_{it}+\tau_r r_{it})\right),\\
C_g&=\left\|\frac{1}{|\mathcal{I}_g|}\sum_{(i,t)\in\mathcal{I}_g}u_{it}\right\|_2.
\end{aligned}
\end{equation}
Here $\mathbf{f}_g$ is the local motion field in scaled coordinate space, $A_g$ is normalized activation, and $C_g$ is directional consistency within the cell. The operator $\operatorname{scale}_{[0,1]}(\cdot)$ denotes min--max normalization over cells in the same route, and $\tau_r$ controls the contribution of the second-difference magnitude. In \Cref{fig:geoais_diagnostics}, the speed, acceleration, and progress traces correspond to $m_t$, $r_t$, and $q_t$ in normalized coordinate space.

The encoder-side components of the Maritime Domain Feature Layer operate within the AIS Spatiotemporal Encoder. Vessel Motion Embedding is projected into the convolutional channels through GeoAIS Feature Enhancement, while Hidden Traffic Knowledge Extraction adds a batch-level context residual formed from the hidden-state mean and standard deviation. For a mini-batch $\mathcal{B}$, the encoder is
\begin{equation}
\label{eq:gavae_encoder}
\begin{aligned}
r_i
&=\mathrm{vec}\!\left(
C_L\circ\cdots\circ C_2\left(C_1(x_i)+\alpha_{\varphi}\Pi_{\varphi}(\varphi_i)\right)\right),\\
u_i&=\rho_2\!\left(W_2\rho_1(W_1 r_i+b_1)+b_2\right),\\
\bar{u}_{\mathcal{B}}&=\frac{1}{|\mathcal{B}|}\sum_{j\in\mathcal{B}}u_j,\quad
s_{\mathcal{B}}=\left(\frac{1}{|\mathcal{B}|}\sum_{j\in\mathcal{B}}(u_j-\bar{u}_{\mathcal{B}})^2\right)^{1/2},\\
h_i&=u_i+\alpha_c f_c\!\left([u_i,\bar{u}_{\mathcal{B}},s_{\mathcal{B}}]\right),\quad
\mu_i=W_\mu h_i+b_\mu,\quad
\ell_i=W_\ell h_i+b_\ell.
\end{aligned}
\end{equation}
Here $C_l$ denotes the one-dimensional AIS spatiotemporal encoding blocks, $\Pi_{\varphi}$ projects the Vessel Motion Embedding for GeoAIS Feature Enhancement, $\rho_1$ and $\rho_2$ are nonlinear dense transformations, and $f_c$ performs Hidden Traffic Knowledge Extraction from the batch-level statistics. The resulting Maritime-Aware Latent Space represents route motion and latent maritime regularities in the batch-level route distribution. The posterior is Gaussian, with log-variance vector $\ell_i$, and the latent code is sampled by reparameterization \cite{kingma2013autoencoding}:
\begin{equation}
\label{eq:gavae_posterior}
q_\phi(z_i|x_i)=\mathcal{N}\!\left(\mu_i,\mathrm{diag}(\exp(\ell_i))\right),
\quad
z_i=\mu_i+\exp(\ell_i/2)\odot\epsilon_i,\quad
\epsilon_i\sim\mathcal{N}(0,I).
\end{equation}
For generation, an anchor index is drawn uniformly with replacement from the processed training trajectories and the latent code is sampled from its approximate posterior. The latent distribution is the empirical aggregate posterior
\begin{equation}
\hat{q}_{\phi}(z)=\frac{1}{N_r}\sum_{i=1}^{N_r}q_{\phi}(z\mid x_i).
\end{equation}
The standard normal distribution serves as the reference distribution in the KL term. Generation therefore produces route-conditioned trajectory variations from the empirical aggregate posterior.

The AIS Spatiotemporal Decoder reconstructs each trajectory through a direct coordinate path and a cumulative motion path. The cumulative motion path and learned gate form GeoAIS Feature Alignment, which blends the two reconstructions. Given decoder features $r_t(z)$, the two paths are
\begin{equation}
\label{eq:gavae_decoder}
\begin{aligned}
\tilde{p}^{\mathrm{coord}}_t&=\operatorname{sigmoid}\!\left(D_t(r(z))\right),\\
p^{0}(z)&=\operatorname{sigmoid}(W_a z+b_a),\quad
\Delta\tilde{p}_t=0.05\,\tanh(W_{\Delta}r_t(z)+b_{\Delta}),\\
\tilde{p}^{\mathrm{mot}}_t&=\operatorname{clip}\!\left(p^{0}(z)+\sum_{k=1}^{t}\Delta\tilde{p}_k,0,1\right),\\
\gamma_t&=\operatorname{sigmoid}(W_{\gamma}r_t(z)+b_{\gamma}),\\
\hat{p}_t&=\gamma_t\odot\tilde{p}^{\mathrm{coord}}_t+(1-\gamma_t)\odot\tilde{p}^{\mathrm{mot}}_t.
\end{aligned}
\end{equation}
The direct path reconstructs coordinate patterns along the route, while the cumulative motion path represents an anchored sequence of vessel displacements. The GeoAIS Feature Alignment gate determines the contribution of each path at every time step and coordinate dimension.

\begin{figure}[!ht]
\centering
\includegraphics[width=0.65\linewidth]{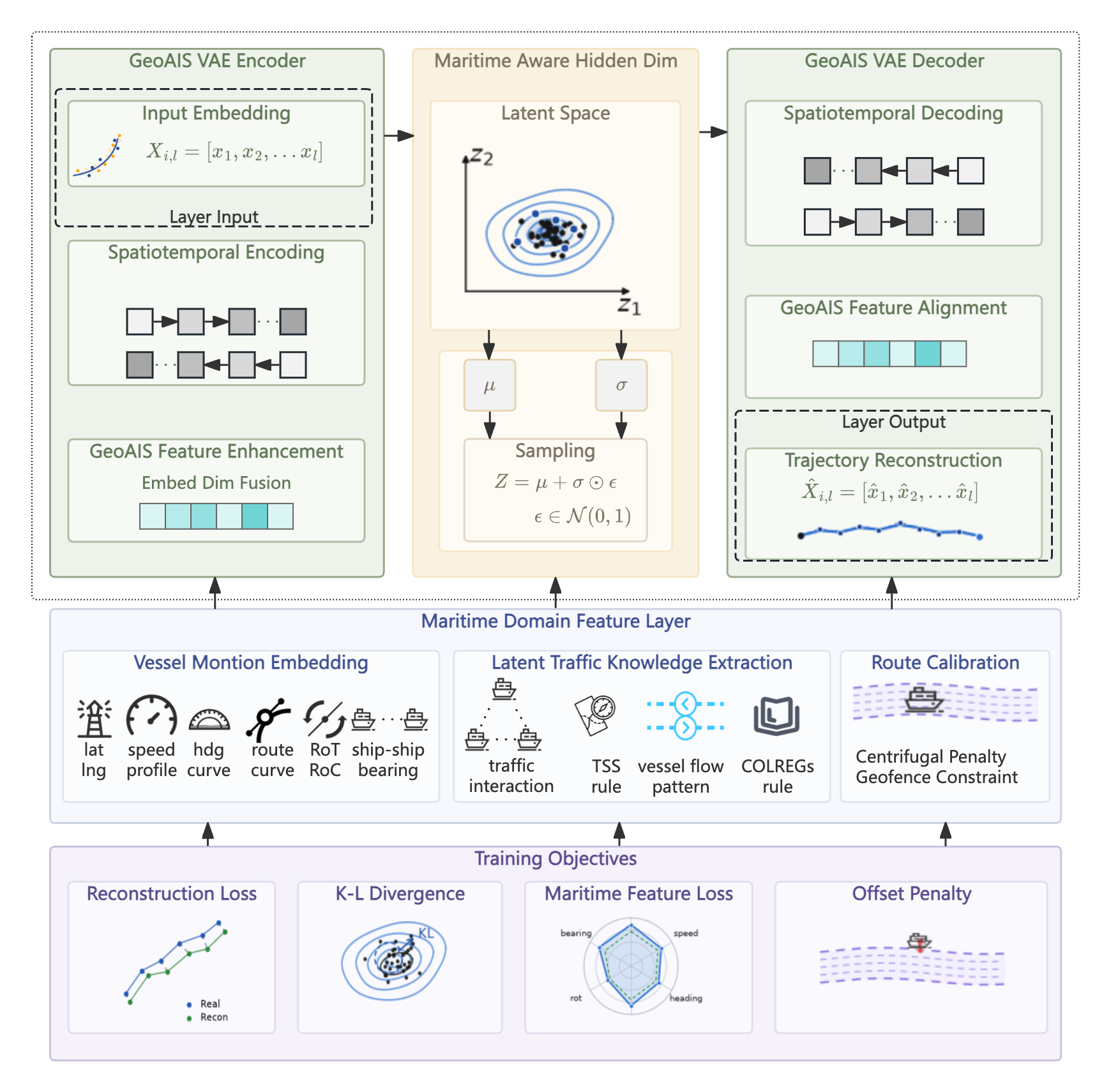}
\caption{GeoAIS VAE architecture and Maritime Domain Feature Layer.}
\label{fig:gavae_structure}
\end{figure}

After inverse scaling, Route Calibration aligns the generated trajectories with the observed route means at each time step and the global coordinate spread:
\begin{equation}
\label{eq:route_calibration}
\hat{X}^{\mathrm{cal}}
=(1-\rho)\hat{X}
{}+\rho\left[
\frac{\hat{X}-\mu_t(\hat{X})}{\sigma_g(\hat{X})+\epsilon_c}\odot\sigma_g(X)+\mu_t(X)
\right],
\end{equation}
where $\mu_t(\cdot)$ is the route mean across training samples at each time step and $\sigma_g(\cdot)$ is the global coordinate standard deviation over the training samples and time steps. The trajectories after Route Calibration form the Generated Trajectory Pools used in scenario construction, and \Cref{fig:gavae_structure} summarizes the GAVAE architecture.

The training objective combines Reconstruction Loss, KL Divergence, Maritime Feature Loss, and Offset Penalty:
\begin{equation}
\label{eq:gavae_loss}
\mathcal{L}
=\mathcal{L}_{\mathrm{rec}}
{}+\beta\mathcal{L}_{\mathrm{KL}}
{}+\mathcal{L}_{\mathrm{mar}}
{}+\lambda_{\mathrm{off}}\mathcal{L}_{\mathrm{off}}.
\end{equation}
The Reconstruction Loss preserves coordinate and marginal trajectory statistics:
\begin{equation}
\label{eq:gavae_reconstruction_loss}
\begin{aligned}
\mathcal{L}_{\mathrm{rec}}
&=\sum_{i,t,d}(x_{itd}-\hat{x}_{itd})^2
{}+\frac{1}{ND}\sum_i\left\|\frac{1}{T}\sum_t x_{it\cdot}-\frac{1}{T}\sum_t\hat{x}_{it\cdot}\right\|_2^2\\
&\quad {}+\frac{1}{NT}\sum_i\left\|\frac{1}{D}\sum_d x_{i\cdot d}-\frac{1}{D}\sum_d\hat{x}_{i\cdot d}\right\|_2^2.
\end{aligned}
\end{equation}
The Maritime Feature Loss emphasizes lower-route and edge-region reconstruction:
\begin{equation}
\label{eq:gavae_maritime_feature_loss}
\mathcal{L}_{\mathrm{mar}}
=\sum_{i,t,d}\left[
\lambda_{\mathrm{low}}\mathbf{1}\{x_{it,\mathrm{lat}}\leq 0.5\}
{}+\lambda_{\mathrm{edge}}\frac{1}{D}\sum_{d'=1}^{D}2|x_{itd'}-0.5|
\right](x_{itd}-\hat{x}_{itd})^2.
\end{equation}
The KL Divergence regularizes each approximate posterior toward the standard normal reference distribution used in the training objective:
\begin{equation}
\mathcal{L}_{\mathrm{KL}}
=-\frac{1}{2N}\sum_i\sum_{k=1}^{K}\left(1+\ell_{ik}-\mu_{ik}^2-\exp(\ell_{ik})\right).
\end{equation}
The Offset Penalty compares generated and AIS route-envelope statistics through spread at each time step and moments in the lower corridor:
\begin{equation}
\label{eq:gavae_offset_penalty}
\begin{aligned}
\mathcal{L}_{\mathrm{off}}
&=N\|S_t(X)-S_t(\hat{X})\|_F^2
{}+0.1NTD\|M_L(X)-M_L(\hat{X};X)\|_2^2\\
&\quad {}+NTD\|S_L(X)-S_L(\hat{X};X)\|_2^2.
\end{aligned}
\end{equation}
where $S_t(\cdot)$ is the coordinate standard deviation at each time step, and $M_L(\cdot)$ and $S_L(\cdot)$ are the weighted mean and standard deviation in the lower corridor.
Vessel Motion Embedding and GeoAIS Feature Alignment represent sequential coordinate variation. A Savitzky--Golay filter smooths the generated trajectories used in visualization and scenario construction; quantitative benchmark tables report the generated trajectories before filtering.

Both route models use common implementation settings. The AIS Spatiotemporal Encoder uses five one-dimensional convolution blocks with 64 channels and kernel sizes 10, 2, 2, 2, and 4, followed by 512- and 64-dimensional dense layers that parameterize a 100-dimensional latent distribution. The AIS Spatiotemporal Decoder maps the latent code through 64-, 512-, and 64-dimensional dense layers before GeoAIS Feature Alignment blends a transposed-convolution coordinate path and a cumulative motion path. GAVAE is trained for 2000 epochs with Adam using a learning rate of 0.001, $\epsilon=10^{-7}$, batch size 770, and dropout 0.1; validation loss determines the reported checkpoint. The objective uses $\beta=1.0$ for KL Divergence, an Offset Penalty weight of 1.0, a lower-route Maritime Feature Loss weight of 0.75, and an edge-region weight of 0.35. After sampling from the training aggregate posterior, 1000 generated trajectories are inverse-scaled and processed by Route Calibration with blend factor 0.9. The calibration ablation evaluates the same decoded trajectories before this transformation. The scenario-construction experiment uses a deterministic prefix of 192 trajectories from each generated trajectory pool.

\subsection{Trajectory-Generation Metrics}
\label{subsec:trajectory_generation_metrics}

Trajectory generation quality is evaluated on the vessel-disjoint test sets with both pointwise and route-scale metrics. Mean absolute error (MAE) and mean squared error (MSE) average coordinate errors over all generated and real trajectory combinations, following vessel trajectory generation studies \cite{murray2020dual}. Summary distance (SD) compares global and coordinate-wise trajectory extrema. Distance metric (DM) measures the Gaussian Fréchet distance between normalized 20-dimensional trajectory summary descriptor distributions. Maximum mean discrepancy (MMD) compares flattened generated and real trajectory distributions with a unit-bandwidth RBF kernel two-sample statistic \cite{gretton2012kernel}. Dynamic time warping (DTW) measures temporal shape similarity on deterministic trajectory samples and is averaged over both nearest-neighbor directions \cite{sakoe1978dynamic}. Bbox coverage (BC) measures route envelope coverage using the ratio of generated and real bounding box areas. Lower values are better for MAE, MSE, SD, DM, MMD, and DTW; higher BC indicates broader spatial coverage.

\subsection{Encounter Pairing and Safety-Critical Scenario Construction}
Let $\hat{\mathcal{T}}^{(1)}$ and $\hat{\mathcal{T}}^{(2)}$ denote the generated trajectory pools for the two traffic flows. Crossing and head-on candidates pair trajectories from different flows, while overtaking candidates pair two different trajectories from the same flow:
\begin{equation}
\label{eq:scenario_operators}
\begin{aligned}
\mathcal{P}_{C}&=\hat{\mathcal{T}}^{(1)}\times\hat{\mathcal{T}}^{(2)},\\
\mathcal{P}_{H}&=\hat{\mathcal{T}}^{(1)}\times\hat{\mathcal{T}}^{(2)},\\
\mathcal{P}_{O}^{(r)}&=\{(\tau_i^{(r)},\tau_j^{(r)}):\tau_i^{(r)},\tau_j^{(r)}\in\hat{\mathcal{T}}^{(r)},i\neq j\},
\quad r\in\{1,\ldots,R\}.
\end{aligned}
\end{equation}
Thus, overtaking construction can be instantiated on any route pool by selecting two different trajectories from that same pool; it does not depend on a particular route. Each ordered pair is evaluated under temporal offsets $\delta\in\Delta$. A type-specific relative-motion consistency predicate separates crossing, head-on, and overtaking candidates after alignment.

For each candidate type $c$, the encounter region $\mathcal{R}_c$ is the intersection of the axis-aligned spatial envelopes of its two trajectory pools. Both vessel states must lie in $\mathcal{R}_c$ when the dynamic encounter condition is evaluated. This data-derived overlap region excludes apparent interactions outside the shared traffic area.

The temporally aligned trajectories are projected into a local nautical-mile frame:
\begin{equation}
\label{eq:nm_projection}
\Pi_{\mathrm{nm}}(p_t)=
\begin{bmatrix}
60\cos(\pi\phi_c/180)(\lambda_t-\lambda_c)\\
60(\phi_t-\phi_c)
\end{bmatrix},
\quad
(\lambda_c,\phi_c)=\frac{1}{|\mathcal{P}|}\sum_{p\in\mathcal{P}}p,
\end{equation}
where longitude and latitude are in degrees and $\mathcal{P}$ contains the points in the two trajectory pools used for the current candidate type. Let $a_k$ and $b_k$ denote an aligned trajectory pair, and define the relative position and velocity as $r_k=b_k-a_k$ and $v_k=\dot{b}_k-\dot{a}_k$. The observed separation and dynamic CPA indicators are
\begin{equation}
\label{eq:dynamic_cpa}
\begin{aligned}
D_r(k)&=\|r_k\|_2,\\
\mathrm{TCPA}(k)&=-\frac{r_k^\top v_k}{\|v_k\|_2^2},\\
\mathrm{DCPA}(k)&=\|r_k+\mathrm{TCPA}(k)v_k\|_2.
\end{aligned}
\end{equation}
The closest-interaction index and minimum observed separation are
\begin{equation}
\label{eq:closest_approach}
k^\star=\arg\min_k D_r(k),\qquad D_{\min}=D_r(k^\star).
\end{equation}
An aligned pair is retained only if $D_{\min}\leq d_{\min}$ and there exists an index $k$ at which both states lie in $\mathcal{R}_c$ and
\begin{equation}
\label{eq:safety_critical_filter}
D_r(k)\leq d_{\mathrm{th}},\qquad
0<\mathrm{TCPA}(k)\leq T_{\mathrm{th}},\qquad
\mathrm{DCPA}(k)\leq d_{\mathrm{cpa}}.
\end{equation}
The type-specific relative-motion condition and the availability of a complete temporal window around $k^\star$ are also required. For a retained pair, the reported dynamic DCPA and associated TCPA are taken at the admissible index with the minimum DCPA.

\subsection{Configurable Safety-Critical Scenario Representation}
Given a retained trajectory pair $(\tau_i,\tau_j)$ that satisfies the safety-critical encounter conditions, a standardized scenario is constructed around the closest-interaction instant. Let $t^\star$ be the source-trajectory time corresponding to $k^\star$ after applying the temporal offset.

Two temporal margins, $t_{\mathrm{early}}$ and $t_{\mathrm{after}}$, define the encounter interval
\begin{equation}
\mathcal{I}_{\mathrm{enc}}=
[t^\star-t_{\mathrm{early}},\,t^\star+t_{\mathrm{after}}].
\end{equation}
Accordingly, each vessel trajectory is partitioned into three segments:
\begin{enumerate}
    \item \textbf{Pre-encounter segment (nominal tracking path):} from the trajectory start to $t^\star-t_{\mathrm{early}}$. This segment provides the nominal reference path before the vessel enters the safety-critical encounter interval.
    \item \textbf{Encounter segment (safety-critical test window):} from $t^\star-t_{\mathrm{early}}$ through $t^\star+t_{\mathrm{after}}$, including both interval endpoints. This segment defines the interval in which collision-avoidance and maneuvering performance are evaluated.
    \item \textbf{Post-encounter segment (recovery tracking path):} after $t^\star+t_{\mathrm{after}}$ to the trajectory end. This segment provides the nominal recovery path after the safety-critical interaction.
\end{enumerate}

The extracted scenario instance is represented as
\begin{equation}
\label{eq:scenario_segment}
\sigma_{ij}=\left(i,j,t^\star,t_{\mathrm{early}},t_{\mathrm{after}}\right).
\end{equation}
together with the corresponding pre-encounter, encounter, and post-encounter trajectory clips for both vessels. This representation provides a common interface for simulation-based digital testing: the pre-encounter segment initializes nominal tracking, the encounter segment defines the collision-avoidance test window, and the post-encounter segment supports recovery evaluation.

\section{Experimental Results}
\subsection{Vessel Trajectory Generation for Designated Traffic Flows}
\subsubsection{Data Description}
\begin{figure}[!htbp]
\centering
\begin{minipage}[b]{0.49\linewidth}
    \centering
    \includegraphics[width=\linewidth]{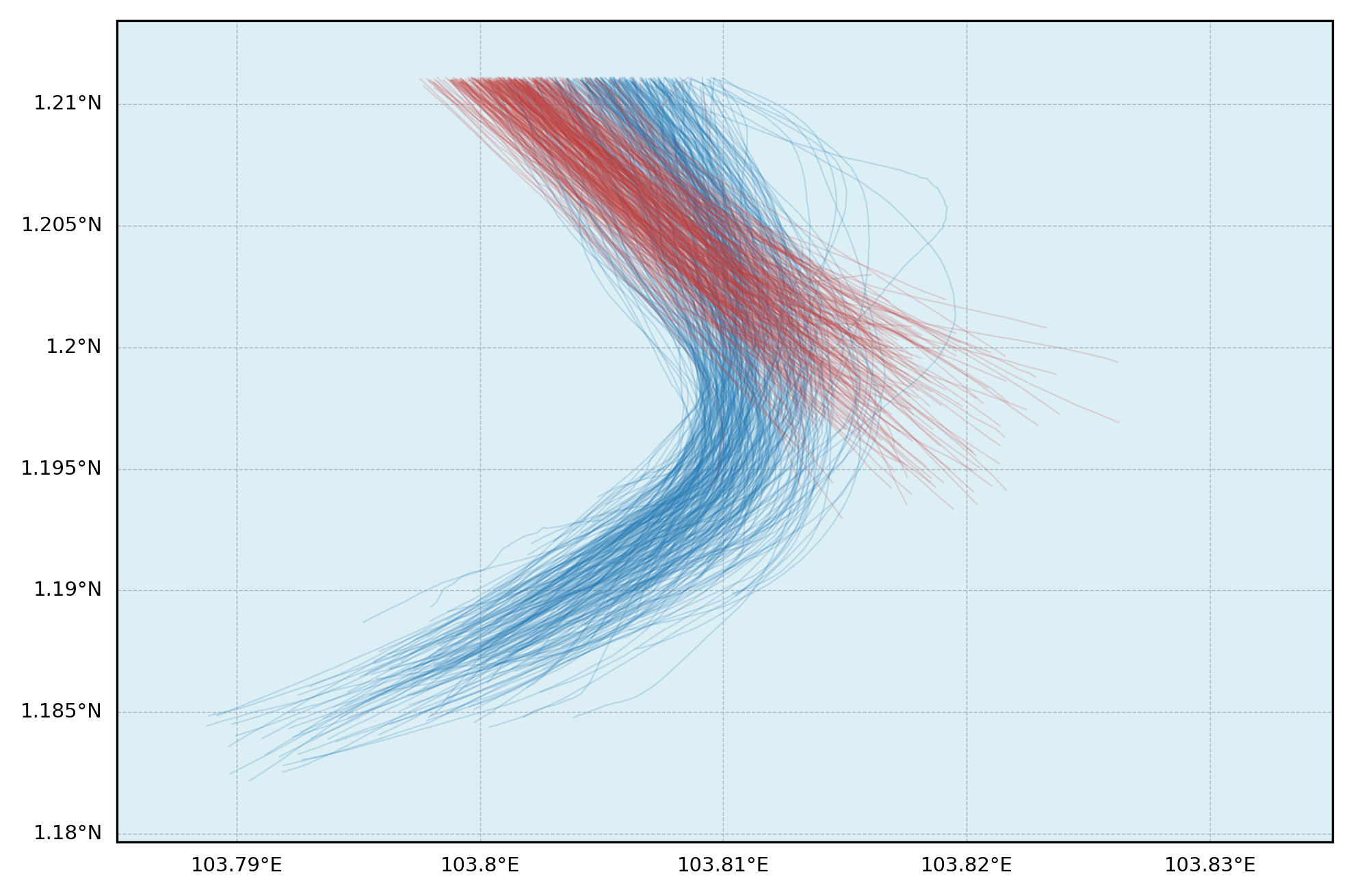}
    (a) AIS trajectories
\end{minipage}
\hfill
\begin{minipage}[b]{0.49\linewidth}
    \centering
    \includegraphics[width=\linewidth]{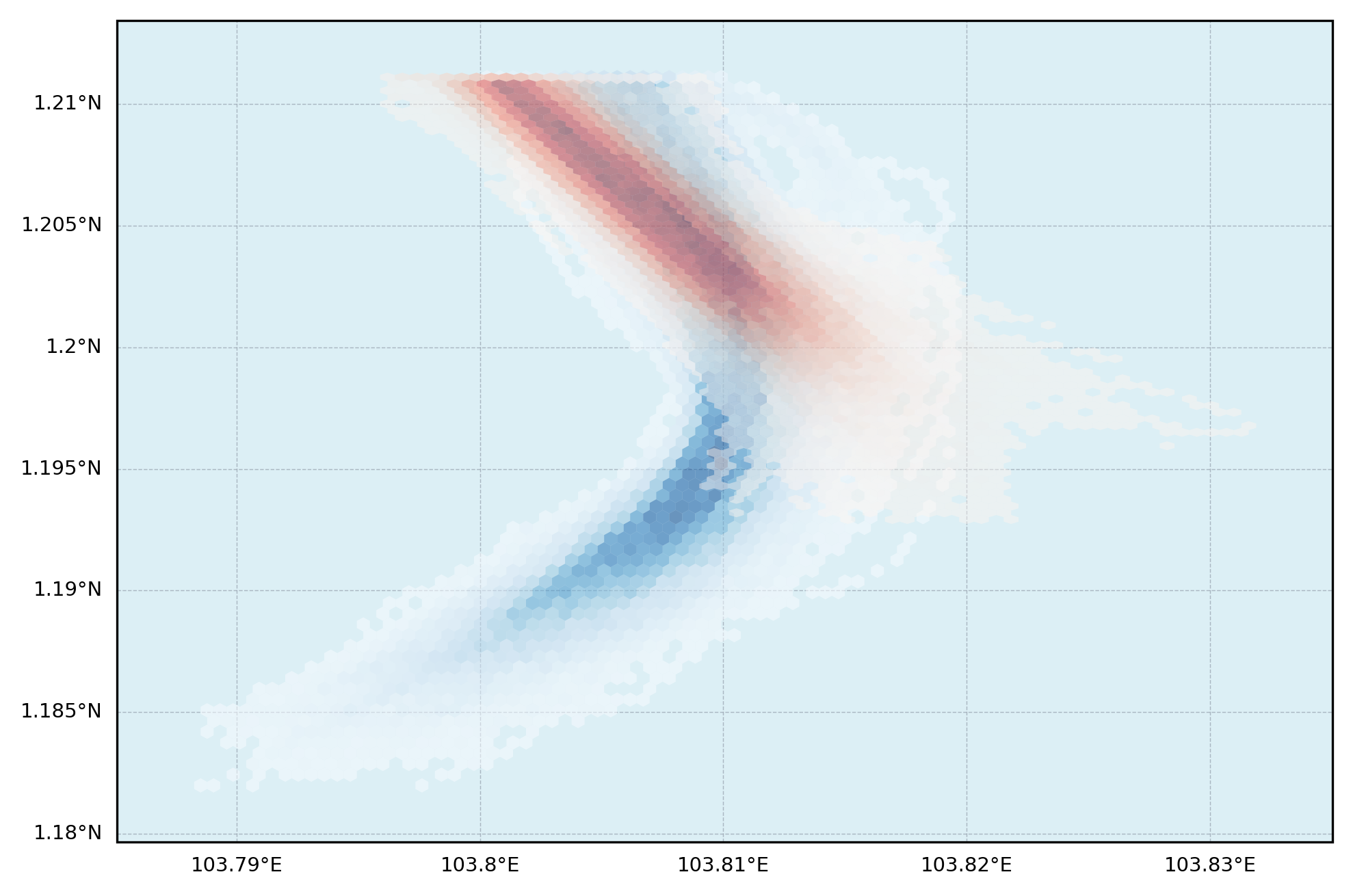}
    (b) Sample density
\end{minipage}
\caption{Study Area of Route 1\&2 AIS trajectories with sample density in the Singapore Strait.}
\label{fig:route_visualization}
\end{figure}

The dataset contains one year of AIS records from the Singapore Strait within 103.785°E--103.837°E and 1.180°N--1.215°N. Two opposing traffic flows in the same corridor define Route 1 (northbound) and Route 2 (southbound), creating spatially overlapping trajectories and potential encounters. Median vessel length and speed are 95~m and 6.59~kn for Route 1, and 183~m and 6.23~kn for Route 2. Trajectories are selected by geographic bounds, resampled at 10-second intervals, interpolated, filtered to retain the dominant flow, and standardized using fixed time windows. This yields 1,094 Route 1 trajectories with 91 steps and 2,310 Route 2 trajectories with 61 steps. Each route is split by MMSI into separate training, validation, and test sets, avoiding seperation of certain vessel into different sets. The training set determines normalization and Route Calibration statistics, the validation set selects checkpoints, and the test set evaluates trajectory generation. \Cref{fig:route_visualization} shows the trajectory distribution and sample density.

\subsubsection{Experiment Setup}
The proposed method and comparison models are trained on Ubuntu 22.04 with a single RTX 4090 under PyTorch 2.6. All models are trained for 2000 epochs and generate 1000 trajectories per route. Scenario construction uses the same deterministic prefix of 192 trajectories from each generated pool and temporal offsets from $-300$~s to $300$~s at 30~s intervals. The matching thresholds are $d_{\min}=0.05$~nm, $d_{\mathrm{th}}=0.50$~nm, $T_{\mathrm{th}}=600$~s, and $d_{\mathrm{cpa}}=0.50$~nm, with $t_{\mathrm{early}}=t_{\mathrm{after}}=100$~s.

\subsubsection{Comparison Studies}
Route 1\&2 are benchmarked as the two datasets for trajectory generation, training, testing, and evaluation were performed independently.  CNN \cite{lecun1998gradientbased} represents convolutional reconstruction, while GAN \cite{goodfellow2014generative} represents adversarial generation. Base VAE \cite{kingma2013autoencoding}, LSTM \cite{hochreiter1997long}, and BiLSTM \cite{schuster1997bidirectional} represent variational and recurrent sequence models. SocialVAE \cite{xu2022socialvae} introduces social-context trajectory generation. KoVAE \cite{naiman2024generative}, TC-VAE \cite{acciaio2026timecausal}, and TARFVAE \cite{wei2025tarfvae} represent recent VAE-based time-series generators.

\Cref{tab:benchmark_combined} compares the generated trajectories before filtering on the vessel-disjoint test sets using the metrics defined in \Cref{subsec:trajectory_generation_metrics}.
GAVAE with Route Calibration ranks first in SD, MMD, DTW, and BC on both routes and second in DM; its bounding-box coverage reaches 0.980 on Route 1 and 0.998 on Route 2, supporting the generated trajectory pools used for scenario construction.

\begin{table}[!htbp]
\centering
\caption{Trajectory-generation results on Route 1 and Route 2}
\label{tab:benchmark_combined}
\resizebox{\linewidth}{!}{%
\begin{tabular}{@{}lc*{14}{r}@{}}
\toprule
\multirow{2}{*}{Model}
& \multirow{2}{*}{Year}
& \multicolumn{7}{c}{Route 1}
& \multicolumn{7}{c}{Route 2} \\
\cmidrule(lr){3-9}
\cmidrule(lr){10-16}
&
& MAE\textcolor{red}{$\downarrow$}
& MSE\textcolor{red}{$\downarrow$}
& SD\textcolor{blue}{$\downarrow$}
& DM\textcolor{blue}{$\downarrow$}
& MMD\textcolor{blue}{$\downarrow$}
& DTW\textcolor{blue}{$\downarrow$}
& BC\textcolor{blue}{$\uparrow$}
& MAE\textcolor{red}{$\downarrow$}
& MSE\textcolor{red}{$\downarrow$}
& SD\textcolor{blue}{$\downarrow$}
& DM\textcolor{blue}{$\downarrow$}
& MMD\textcolor{blue}{$\downarrow$}
& DTW\textcolor{blue}{$\downarrow$}
& BC\textcolor{blue}{$\uparrow$} \\
\midrule
LSTM
& 1997
& 2.569 & 11.722 & 6.594 & 3.248 & 0.855 & 7.210 & 0.706
& 1.960 & 7.330 & 12.672 & 3.823 & 0.335 & 4.092 & 0.611 \\

BiLSTM
& 1997
& 2.529 & 11.382 & 6.934 & 3.557 & 0.988 & 7.491 & 0.693
& 1.983 & 7.441 & 9.752 & 3.592 & 0.297 & 3.942 & 0.691 \\

CNN
& 1998
& 2.590 & 11.864 & \underline{3.244} & 3.076 & 1.793 & \underline{7.193} & \underline{0.861}
& 2.039 & 7.802 & \underline{4.272} & \textbf{3.131} & 0.880 & 3.785 & \underline{0.867} \\

VAE
& 2013
& 2.298 & 9.581 & 7.597 & \textbf{2.781} & 3.604 & 7.381 & 0.710
& 1.804 & 6.153 & 13.737 & 3.409 & \underline{0.023} & 4.079 & 0.618 \\

GAN
& 2014
& 2.688 & 12.685 & 5.994 & 4.232 & 39.695 & 14.857 & 0.785
& 2.162 & 8.037 & 12.091 & 5.004 & 28.308 & 10.397 & 0.651 \\

ConvVAE
& 2015
& 2.326 & 9.636 & 6.774 & 4.004 & 10.403 & 7.623 & 0.743
& 2.004 & 7.193 & 6.315 & 3.582 & 6.184 & 4.751 & 0.802 \\

Social VAE
& 2022
& \textbf{1.846} & \textbf{5.996} & 19.678 & 6.233 & \underline{0.534} & 12.802 & 0.239
& 1.801 & 6.112 & 17.506 & 4.296 & 0.082 & 5.737 & 0.482 \\

KoVAE
& 2024
& 2.193 & 8.804 & 9.027 & 3.935 & 0.608 & 7.673 & 0.675
& 1.961 & 7.213 & 8.753 & 3.796 & 0.082 & \underline{3.642} & 0.748 \\

TARFVAE
& 2025
& \underline{2.039} & \underline{7.553} & 11.879 & 3.909 & 0.655 & 9.818 & 0.553
& \textbf{1.710} & \underline{5.671} & 15.887 & 4.206 & 0.627 & 4.785 & 0.472 \\

TC-VAE
& 2026
& 2.148 & 8.053 & 8.875 & 5.026 & 1.299 & 9.647 & 0.676
& \underline{1.729} & \textbf{5.556} & 18.838 & 3.250 & 0.090 & 4.389 & 0.582 \\

\midrule
GAVAE
& (Ours)
& 2.461 & 10.946 & \textbf{1.441} & \underline{2.966} & \textbf{0.019} & \textbf{6.667} & \textbf{0.980}
& 1.994 & 7.568 & \textbf{1.181} & \underline{3.133} & \textbf{0.021} & \textbf{3.526} & \textbf{0.998} \\
\bottomrule
\end{tabular}%
}

\par\smallskip
\footnotesize
\textcolor{red}{$\downarrow$}: pairwise metrics;
\textcolor{blue}{$\downarrow$}/\textcolor{blue}{$\uparrow$}: maritime domain metrics;
\textbf{bold}: best numerical result;
\underline{underline}: second-best numerical result.

\par\smallskip
\footnotesize
Scaled values: MAE $10^{-3}$, MSE $10^{-6}$, SD $10^{-3}$,
MMD $10^{-5}$, and DTW $10^{-3}$.
\end{table}

\subsubsection{Ablation Studies}

\Cref{tab:ablation_combined} reports metric-specific trade-offs among Route Calibration, Vessel Motion Embedding, Hidden Traffic Knowledge Extraction, GeoAIS Feature Alignment, Offset Penalty, and Maritime Feature Loss.
Removing Route Calibration increases SD from 1.441 to 3.615 on Route 1 and from 1.181 to 3.077 on Route 2; MMD rises from 0.019 to 1.790 and from 0.021 to 2.085, while BC declines from 0.980 to 0.842 and from 0.998 to 0.888.

\begin{table}[!htbp]
\centering
\caption{GAVAE ablation results on Route 1 and Route 2}
\label{tab:ablation_combined}
\resizebox{\linewidth}{!}{%
\begin{tabular}{@{}lrrrrrrrrrrrrrr@{}}
\toprule
\multirow{2}{*}{Model} & \multicolumn{7}{c}{Route 1} & \multicolumn{7}{c}{Route 2} \\
\cmidrule(lr){2-8}\cmidrule(l){9-15}
& MAE\textcolor{red}{$\downarrow$} & MSE\textcolor{red}{$\downarrow$} & SD\textcolor{blue}{$\downarrow$} & DM\textcolor{blue}{$\downarrow$} & MMD\textcolor{blue}{$\downarrow$} & DTW\textcolor{blue}{$\downarrow$} & BC\textcolor{blue}{$\uparrow$} & MAE\textcolor{red}{$\downarrow$} & MSE\textcolor{red}{$\downarrow$} & SD\textcolor{blue}{$\downarrow$} & DM\textcolor{blue}{$\downarrow$} & MMD\textcolor{blue}{$\downarrow$} & DTW\textcolor{blue}{$\downarrow$} & BC\textcolor{blue}{$\uparrow$} \\
\midrule
GAVAE & 2.461 & 10.946 & \textbf{1.441} & \underline{2.966} & \textbf{0.019} & \textbf{6.667} & \textbf{0.980} & 1.994 & 7.568 & \textbf{1.181} & \textbf{3.133} & \textbf{0.021} & \textbf{3.526} & \textbf{0.998} \\
\midrule
w/o Route Calibration & \underline{2.374} & \textbf{10.114} & 3.615 & 3.283 & 1.790 & 6.706 & 0.842 & 1.972 & \textbf{7.284} & 3.077 & \underline{3.149} & 2.085 & \underline{3.534} & 0.888 \\
w/o Vessel Motion Embedding & \textbf{2.372} & \underline{10.272} & 6.097 & 3.782 & 3.069 & 8.349 & 0.775 & 1.963 & 7.463 & 1.211 & 3.304 & 0.175 & 3.625 & \underline{0.997} \\
w/o Latent Traffic Knowledge & 2.466 & 10.902 & 4.694 & 3.165 & 0.090 & 7.841 & 0.852 & \textbf{1.957} & 7.352 & 2.030 & 3.308 & 0.182 & 3.651 & 0.973 \\
w/o GeoAIS Feature Alignment & 2.447 & 10.798 & \underline{2.072} & 3.058 & \underline{0.037} & 6.926 & \underline{0.915} & 1.985 & 7.605 & \underline{1.194} & 3.307 & 0.245 & 3.692 & \textbf{0.998} \\
w/o Offset Penalty & 2.392 & 10.671 & 4.738 & 3.138 & 0.176 & 8.481 & 0.851 & 1.967 & 7.389 & 3.238 & 3.373 & \underline{0.091} & 3.569 & 0.961 \\
w/o Maritime Feature Loss & 2.502 & 11.294 & 2.552 & \textbf{2.872} & 0.071 & \underline{6.671} & 0.893 & \underline{1.959} & \underline{7.341} & 2.305 & 3.294 & 0.164 & 3.674 & 0.982 \\
\bottomrule
\end{tabular}
}
\end{table}

\subsubsection{Visualization of Generated Results}

\begin{figure}[!htbp]
\centering
\includegraphics[width=.65\linewidth]{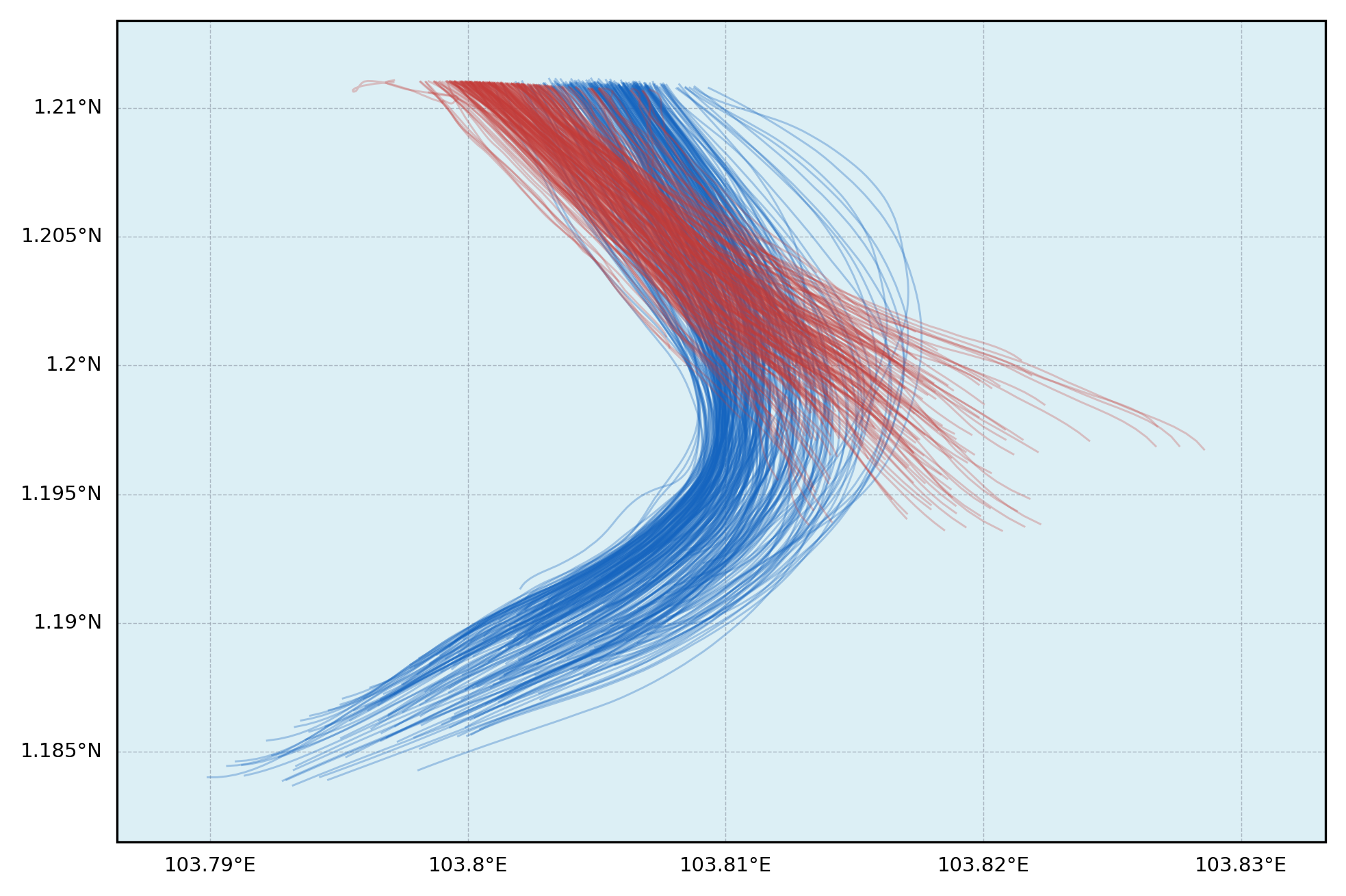}
\caption{Trajectories generated by GAVAE with Route Calibration for Route 1 and Route 2.}
\label{fig:gavae_generated}
\end{figure}

\Cref{fig:gavae_generated} visualizes the generated trajectories for both routes. The trajectories follow the dominant traffic flow envelopes and retain variation within each corridor. The Savitzky--Golay filter is applied after generation for the visualization and scenario construction, while the comparative tables report the generated trajectories before filtering.

\subsection{Demonstration of Constructed Safety-Critical Scenarios}
\subsubsection{Traffic Flow and Encounter Context}
\begin{figure}[!htbp]
\centering

\begin{minipage}[b]{0.49\linewidth}
    \centering
    \includegraphics[width=\linewidth]{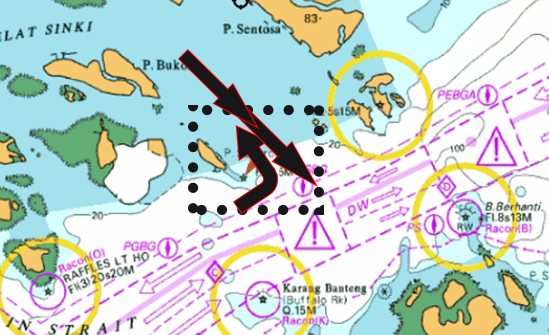}
    
    (a) Study area and traffic flows
    \label{fig:Study_area}
\end{minipage}
\hfill
\begin{minipage}[b]{0.49\linewidth}
    \centering
    \includegraphics[width=\linewidth]{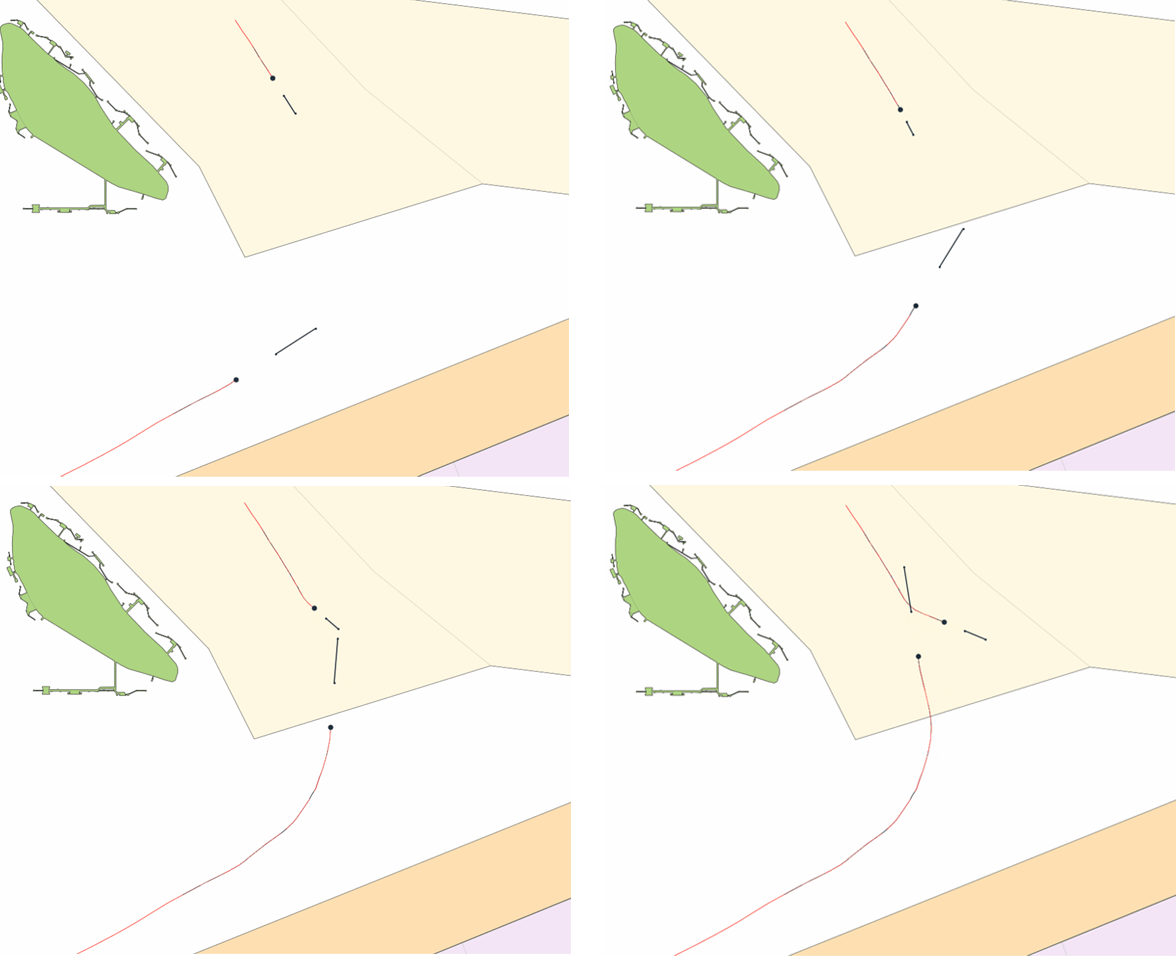}
    (b) Representative encounter example
    \label{fig:contextsub2}
\end{minipage}

\caption{Traffic flow and encounter context in the study area.}
\label{fig:Study_area_example}
\end{figure}

To demonstrate the applicability of the proposed framework, two representative traffic flows within the study area are selected as the experimental contexts. \Cref{fig:Study_area_example}(a), the study area is located near a precautionary area identified on the navigational chart as having a high density of vessel encounters. Among the intersecting traffic flows in this region, the two dominant routes indicated by the thick arrows are selected for analysis: the southbound outbound traffic flow and the eastbound traffic flow that turns northward to enter the port.A representative historical encounter occurring between these traffic flows is illustrated in \Cref{fig:Study_area_example}(b), where the vessel trajectories exhibit a typical interaction pattern observed in the AIS data and the southbound vessel has already initiated an avoidance maneuver. 

\subsubsection{Scenario Construction from Generated Trajectories}
GAVAE provides the generated trajectory pools used in the construction stage. Crossing and head-on candidates pair Route 1 with Route 2. The overtaking operator can be applied to any individual route pool. Each ordered pair is evaluated under the temporal offsets specified in the experiment setup. A candidate is retained only when it satisfies the overlap-region constraint, the minimum observed separation threshold, the dynamic CPA conditions, the encounter-type consistency condition, and the complete pre-encounter and post-encounter margins.

\begin{figure}[!htbp]
\centering
\begin{minipage}[b]{0.32\linewidth}
    \centering
    \includegraphics[width=\linewidth,height=.15\textheight,keepaspectratio]{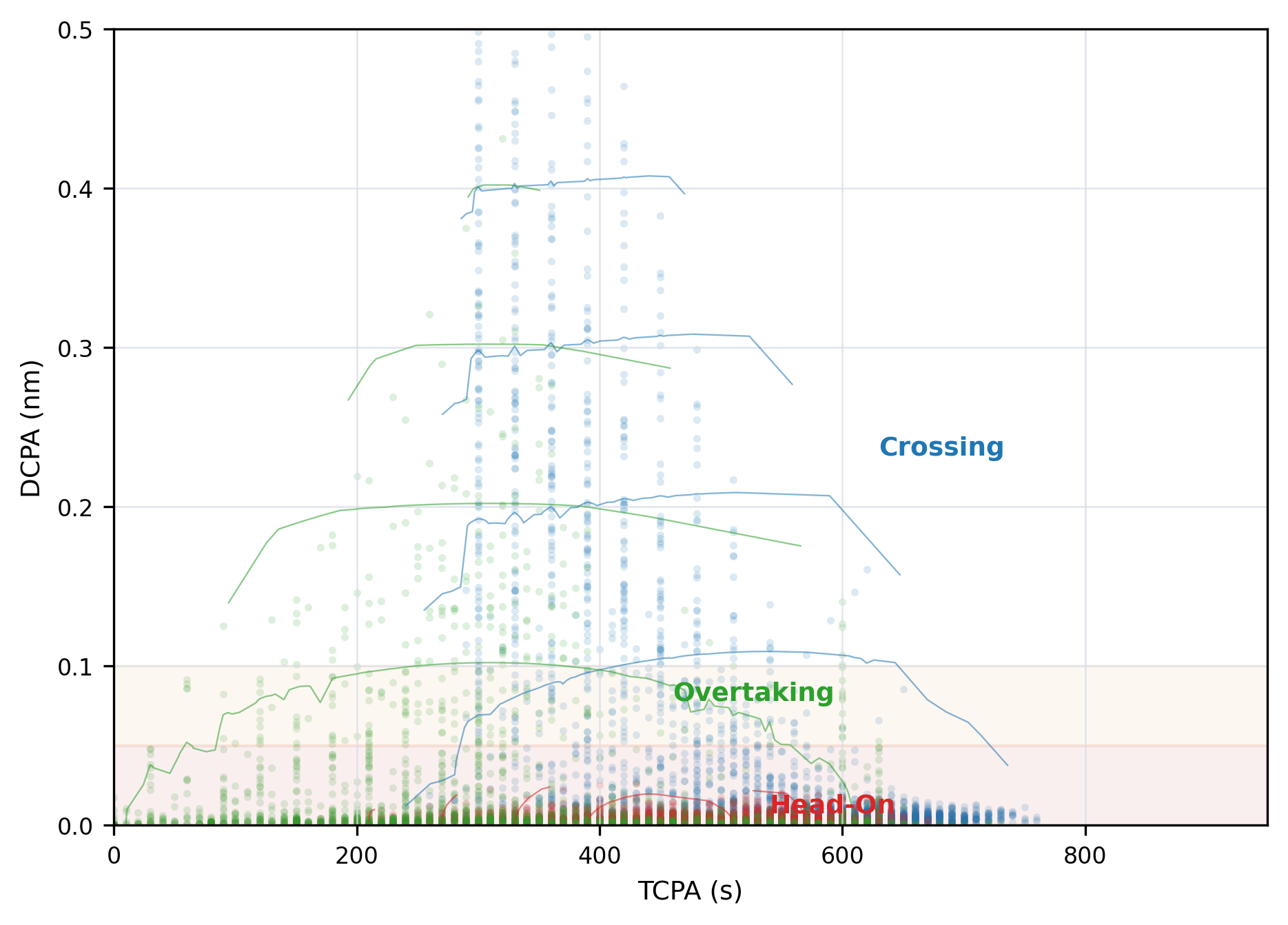}\\
    \textbf{(a)}
\end{minipage}
\hfill
\begin{minipage}[b]{0.32\linewidth}
    \centering
    \includegraphics[width=\linewidth,height=.15\textheight,keepaspectratio]{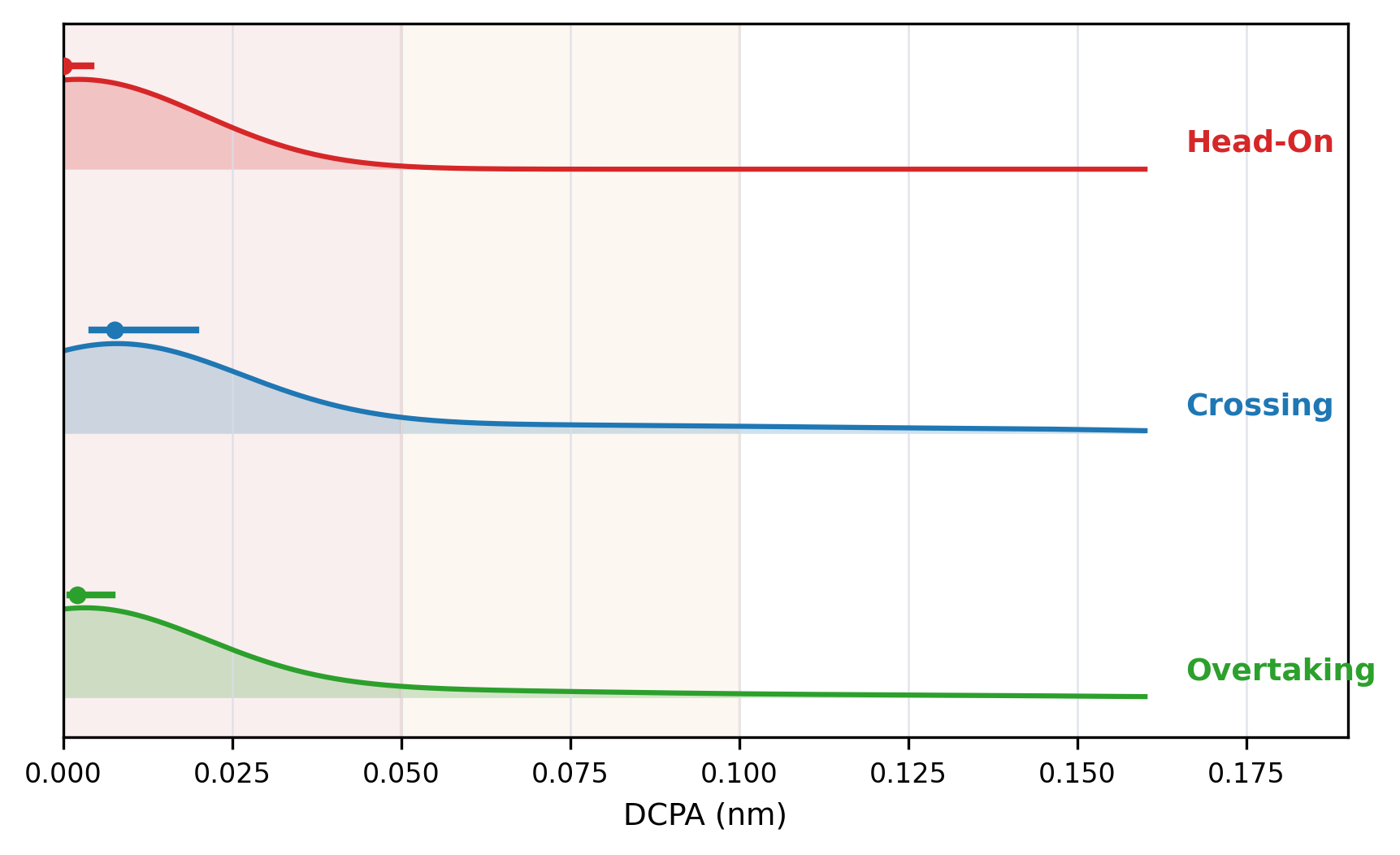}\\
    \textbf{(b)}
\end{minipage}
\hfill
\begin{minipage}[b]{0.32\linewidth}
    \centering
    \includegraphics[width=\linewidth,height=.15\textheight,keepaspectratio]{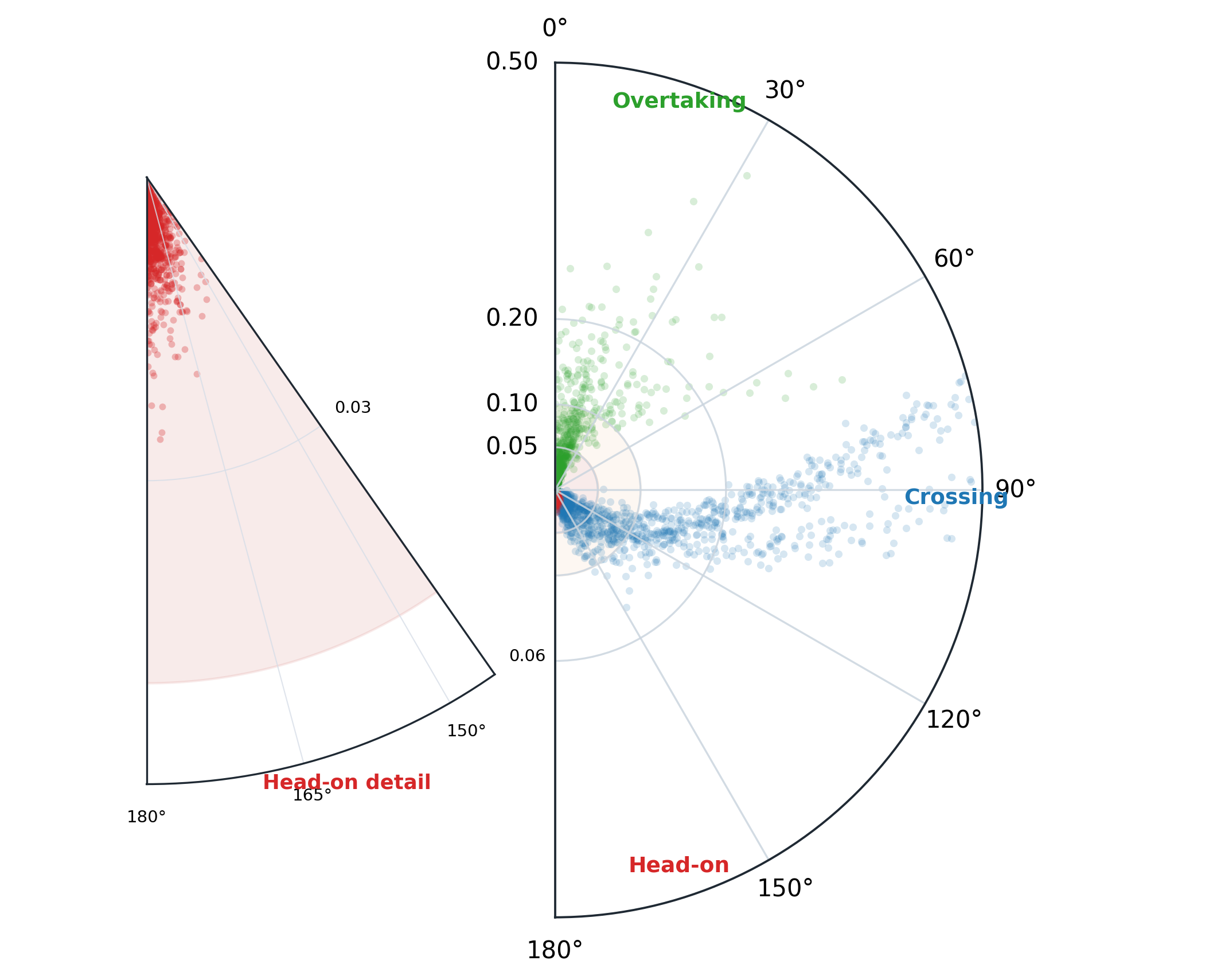}\\
    \textbf{(c)}
\end{minipage}
\caption{DCPA, TCPA, and relative course angle distributions by intended encounter type.}
\label{fig:scenario_risk_summary}
\end{figure}

In \Cref{fig:scenario_risk_summary}, panel (a) shows the joint distribution of DCPA and TCPA for the three intended encounter types. Crossing scenarios cover the widest TCPA range and generally extend to larger DCPA values, while overtaking cases are concentrated at lower DCPA levels. Head-on cases remain strongly clustered near zero DCPA, indicating the highest geometric conflict among the three groups. Panel (b) confirms these differences in the marginal DCPA distributions: head-on encounters are concentrated at the smallest distances, crossing cases have the largest spread, and overtaking cases lie between them. Panel (c) further shows that the encounter types occupy distinct relative-course-angle regions. Crossing cases cluster near $90^\circ$, head-on cases near $180^\circ$, and overtaking cases near small relative angles. These patterns show that the generated scenarios preserve clear geometric differences between the intended encounter types.

\begin{figure}[!htbp]
\centering
\includegraphics[width=\textwidth]{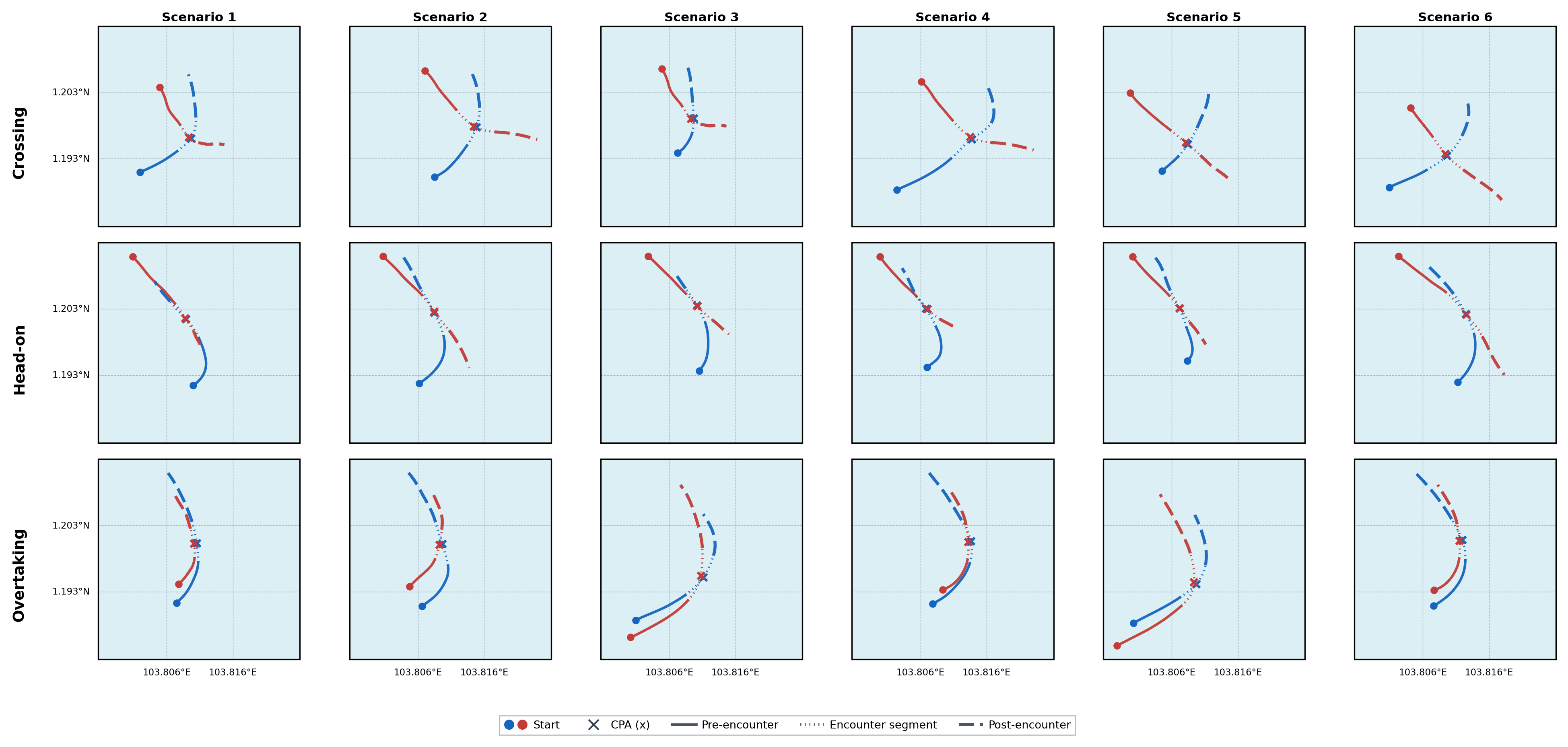}
\caption{Constructed safety-critical encounter scenarios.}
\label{fig:generated}
\end{figure}

\Cref{fig:generated} presents six representative safety-critical encounters selected from the generated scenario pool. The examples show different relative approach geometries while preserving continuous vessel motion across the pre-encounter, encounter, and post-encounter stages. All illustrated cases have nonzero DCPA values close to 0.05~nm, indicating close but non-colliding interactions. The trajectories remain spatially smooth before and after CPA, suggesting that the construction process does not introduce abrupt motion changes. Together, these examples demonstrate that the framework can generate diverse close-encounter configurations while maintaining plausible trajectory continuity and controllable risk levels. The full scenario set covers DCPA values up to 0.50~nm, with $t_{\text{early}}$ and $t_{\text{after}}$ both set to 100~s.

\subsection{Discussion}

The experimental results demonstrate the effectiveness of the proposed framework at both the trajectory-generation and scenario-construction levels. At the individual-trajectory level, GAVAE achieves the best overall performance among the compared generative models. Although MAE and MSE measure pointwise reconstruction accuracy, they do not fully reflect the spatial coverage, route variability, and distributional quality required for safety-critical scenario construction. The stronger SD, MMD, DTW, and BC results therefore indicate that GAVAE better preserves the geometry and variability of historical AIS traffic.

This improvement also provides meaningful downstream benefits. The realistic and diverse trajectory pools generated by GAVAE offer more feasible combinations for encounter construction. Through trajectory pairing and configurable temporal offsets, the framework produces crossing, head-on, and overtaking scenarios with distinct relative course angle, DCPA, and TCPA distributions. These differences confirm that the constructed scenarios preserve the intended relative-motion characteristics rather than representing arbitrary close encounters.

The generated examples further show smooth vessel motion across the pre-encounter, encounter, and post-encounter stages. Meanwhile, the CPA-based thresholds enable the risk level of the retained scenarios to be controlled. This provides the temporal continuity and encounter context needed for simulation-based evaluation of collision avoidance and recovery performance.

Overall, the results establish a clear connection between single-vessel trajectory quality and downstream scenario utility. GAVAE generates realistic and sufficiently diverse trajectories, while the scenario-construction component transforms them into geometrically consistent and configurable safety-critical encounters. Additional representative scenarios are available through the supplementary online visualization.

The current evaluation mainly verifies scenario validity through trajectory-distribution, relative-motion, and CPA-based criteria. These results support the geometric and kinematic plausibility of the scenarios, but do not fully capture interactive vessel behavior.

\section{Conclusion}

This paper presented a data-driven framework for constructing configurable safety-critical vessel encounter scenarios from historical AIS data. The framework combines route-conditioned trajectory generation with trajectory pairing, temporal alignment, encounter-type verification, and CPA-based risk filtering. Experimental results show that GAVAE achieves the strongest overall single-vessel trajectory-generation performance, preserving both route geometry and traffic-flow variability.

The generated trajectory pools support the construction of diverse crossing, head-on, and overtaking scenarios with distinct relative-motion characteristics, controllable risk levels, and continuous vessel motion. Each retained scenario is organized into pre-encounter, encounter, and post-encounter segments, together with vessel states and risk indicators, providing a standardized and simulation-ready representation for MASS testing, benchmarking, and maritime safety assessment. Future work will extend the framework to multi-vessel interactions, incorporate environmental and vessel-dynamics constraints, and support closed-loop simulation and expert validation.

\clearpage
\newpage

\section*{AUTHOR CONTRIBUTIONS}
The authors confirm contribution to the paper as follows:
study conception and design: S.~Sun and L.~Zhao;
data processing and experiments: S.~Sun;
analysis and interpretation of results: S.~Sun, L.~Zhao, and X.~Fu;
draft manuscript preparation: S.~Sun.
All authors reviewed and approved the manuscript.

\section*{DECLARATION OF CONFLICTING INTERESTS}
The authors declared no potential conflicts of interest with respect to the research, authorship, and/or publication of this article.

\section*{FUNDING}
The authors disclosed receipt of the following financial support for the research, authorship, and/or publication of this article: This research was supported by the Singapore Maritime Institute under Maritime Artificial Intelligence Research for Shipping Disruption Evaluation, Launch Boat Optimisation and Digital Testing of Vessel Predictive Maintenance (Grant Reference No. SMI-2025-MTP-02).

\bibliographystyle{trb}
\bibliography{reference}
\end{document}